\documentclass[11pt]{article}
\usepackage{coling2020}
\usepackage{times}
\usepackage{url}
\usepackage{latexsym}

\usepackage{scalerel}
\usepackage{subfigure}

\colingfinalcopy 

\title{Kungfupanda at SemEval-2020 Task 12: BERT-Based Multi-Task Learning for Offensive Language Detection}

\author{Wenliang Dai*, Tiezheng Yu*, Zihan Liu, Pascale Fung \\
Center for Artificial Intelligence Research (CAiRE)\\
Department of Electronic and Computer Engineering\\
The Hong Kong University of Science and Technology, Clear Water Bay, Hong Kong\\
\texttt{\{wdaiai,tyuah,zliucr\}@connect.ust.hk, pascale@ece.ust.hk}}

\date{}

\begin{document}
\maketitle
\begin{abstract}
Nowadays, offensive content in social media has become a serious problem, and automatically detecting offensive language is an essential task. In this paper, we build an offensive language detection system, which combines multi-task learning with BERT-based models.
Using a pre-trained language model such as BERT, we can effectively learn the representations for noisy text in social media.
Besides, to boost the performance of offensive language detection, we leverage the supervision signals from other related tasks.
In the OffensEval-2020 competition, our model achieves 91.51\% F1 score in English Sub-task A, which is comparable to the first place (92.23\% F1). An empirical analysis is provided to explain the effectiveness of our approaches.
\end{abstract}

\section{Introduction}
\label{sec:intro}
%
%
\blfootnote{
    %
    %
    %
    %
    %
    %
    \hspace{-0.65cm}  
    This work is licensed under a Creative Commons 
    Attribution 4.0 International License.
    License details:
    \url{http://creativecommons.org/licenses/by/4.0/}.
}

\blfootnote{
    \hspace{-0.65cm}
    * These two authors contributed equally.
}

Nowadays, offensive content has invaded social media and becomes a serious problem for government organizations, online communities, and social media platforms. Therefore, it is essential to automatically detect and throttle the offensive content before it appears in social media. Previous studies have investigated different aspects of offensive languages such as abusive language~\cite{nobata2016abusive,mubarak2017abusive} and hate speech~\cite{malmasi2017detecting,davidson2017automated}.

Recently, \cite{zampieri2019predicting} first studied the target of the offensive language in twitter and \cite{zampieri-etal-2020-semeval} expand it into the multilingual version, which is practical for studying hate speech concerning a specific target. 
The task is based on a three-level hierarchical annotation schema that encompasses the following three general sub-tasks: (A) Offensive Language Detection; (B) Categorization of Offensive Language; (C) Offensive Language Target Identification.

To tackle this problem, we emphasize that it is crucial to leverage pre-trained language model (e.g., BERT~\cite{devlin2018bert}) to better understand the meaning of sentences and generate expressive word-level representations due to the inherent data noise (e.g., misspelling, grammatical mistakes) in social media (e.g., twitter). In addition, we hypothesize that the internal connections exist among the three general sub-tasks, and to improve one task, we can leverage the information of the other two tasks.
Therefore, we first generate the representations of the input text based on the pre-trained language model BERT, and then we conduct multi-task learning based on the representations.

Experimental results show that leveraging more task information can improve the offensive language detection performance. In the OffensEval-2020 competition, our system achieves 91.51\% macro-F1 score in English Sub-task A (ranked 7th out of 85 submissions). Especially, only the OLID \cite{zampieri2019predicting} is used to train our model and no additional data is used. Our code is available at: \url{https://github.com/wenliangdai/multi-task-offensive-language-detection}.

\section{Related Works}
\label{sec:related_works}
In general, offensive language detection includes some particular types, such as aggression identification~\cite{kumar2018benchmarking}, bullying detection~\cite{huang2014cyber} and hate speech identification~\cite{park2017one}.
\cite{chen2012detecting} applied concepts from NLP to exploit the lexical syntactic feature of sentences for offensive language detection. \cite{huang2014cyber} integrated the textual features with social network features, which significantly improved cyberbullying detection.
\cite{park2017one} and \cite{gamback2017using} used convolutional neural network in the hate-speech detection in tweets.
Recently, \cite{zampieri2019predicting} introduce an offensive language identification dataset, which aims to detect the type and the target of offensive posts in social media. 
\cite{zampieri-etal-2020-semeval} expanded this dataset into the multilingual version, which advances the multilingual research in this area.

Pre-trained language models, such as ELMo~\cite{peters-etal-2018-deep} and BERT~\cite{devlin2018bert} have achieved great performance on a variety of tasks. 
Many recent papers have used a basic recipe of fine-tuning such pre-trained models on a certain domain~\cite{azzouza2019twitterbert,lee2019team,beltagy2019scibert} or on downstream tasks~\cite{howard2018universal,liu2019attention,su2019generalizing}.

\section{Datasets}
\label{sec:datasets}

In this project, two datasets are involved, which are the dataset of OffensEval-2019 and OffensEval-2020 respectively. In this section, we introduce the details of them and discuss our data pre-processing methods. Table \ref{tab:tasks-table} shows the types of labels and how they overlap.

\begin{table}[h]
\centering
\begin{tabular}{|l|c|l|l|l|}
\hline
Task A & \multicolumn{3}{c|}{OFF} & NOT \\ \hline
Task B & \multicolumn{2}{c|}{TIN} & UNT & NULL \\ \hline
Task C & \multicolumn{1}{l|}{IND} & GRP & \multicolumn{2}{c|}{NULL} \\ \hline
\end{tabular}
\caption{Labels of three subtasks.}
\label{tab:tasks-table}
\end{table}

\subsection{Offensive Language Identification Dataset (OLID)}
\label{sec:dataset1}

The OLID \cite{zampieri-etal-2019-predicting} is a hierarchical dataset to identify the type and the target of offensive texts in social media. The dataset is collected on Twitter and publicly available. There are 14,100 tweets in total, in which 13,240 are in the training set, and 860 are in the test set. For each tweet, there are three levels of labels: (A) Offensive/Not-Offensive, (B) Targeted-Insult/Untargeted, (C) Individual/Group/Other. The relationship between them is hierarchical. If a tweet is offensive, it can have a target or no target. If it is offensive to a specific target, the target can be an individual, a group, or some other objects. 
This dataset is used in the OffensEval-2019 competition in SemEval-2019 \cite{zampieri-etal-2019-semeval}. The competition contains three sub-tasks, each corresponds to recognizing one level of label in the dataset.

\subsection{Semi-Supervised Offensive Language Identification Dataset (SOLID)}
\label{sec:dataset2}


A semi-supervised offensive language detection dataset (SOLID) \cite{rosenthal2020} for English is proposed in the OffensEval-2020 competition in SemEval-2020 \cite{zampieri-etal-2020-semeval}. Similar to OLID \cite{zampieri-etal-2019-predicting}, it still has three levels but the data in level A is separated from levels B and C. In level A, there are 9,089,140 tweets, in levels B and C, there are different 188,973 tweets. For each entry of data in level A, the mean and standard deviation of confidence scores generated by the democratic co-training approach are provided as supervision. Because of this, the data is more noisy than OLID. For level B and C, the data is still manually annotated.

\subsection{Data Pre-processing}
\label{sec:data_preprocessing}

Data pre-processing is crucial to this task as the data from Twitter is noisy and sometimes disordered. Moreover, people tend to use more Emojis and hashtags on Twitter, which are unusual in other situations. 

Firstly, all characters are converted to lowercase, and the spaces at ends are stripped. Then, inspired by \cite{zampieri-etal-2019-semeval,liu-etal-2019-nuli}, we further process the dataset in five specific aspects:

\paragraph{Emoji to word.} We convert all emojis to words with corresponding semantic meanings. For example, \scalerel*{\includegraphics[height=\textheight]{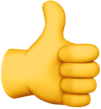}}{B} is converted to \textit{thumbs up}. We achieve this by first utilizing a third-party Python library \footnote{https://github.com/carpedm20/emoji}, and then removing useless punctuation in it.

\paragraph{Hashtag segmentation.} All hashtags in the tweets are segmented by recognizing the capital characters. For example, \textit{\#KeithEllisonAbuse} is transformed to \textit{keith ellison abuse}. This is also achieved by using a third-party Python library \footnote{https://github.com/grantjenks/python-wordsegment}.

\paragraph{User mention replacement.} After reviewing the dataset, we find out that the token \textit{@USER} appears very frequently (a single tweet can have multiple of them), which is a typical phenomenon in tweets. As a result, for those with more than one \textit{@USER} token, we replace all of them with one \textit{@USERS} token. In this way, we remove the redundant words while keeping the key information, which is useful for recognizing targets if there is any.

\paragraph{Rare word substitution.} We substitute some out-of-vocabulary (OOV) words with their synonyms. For example, every \textit{URL} is replaced with a special token, \textit{http}.

\paragraph{Truncation.} We truncate all the tweets to a max length of 64. Although this can get rid of some information in the data, it lowers the GPU memory usage and slightly improves the performance.

\begin{figure*}[t]
    \centering
    \subfigure[Baseline (BERT)]{\label{fig:baseline}\includegraphics[width=.45\linewidth]{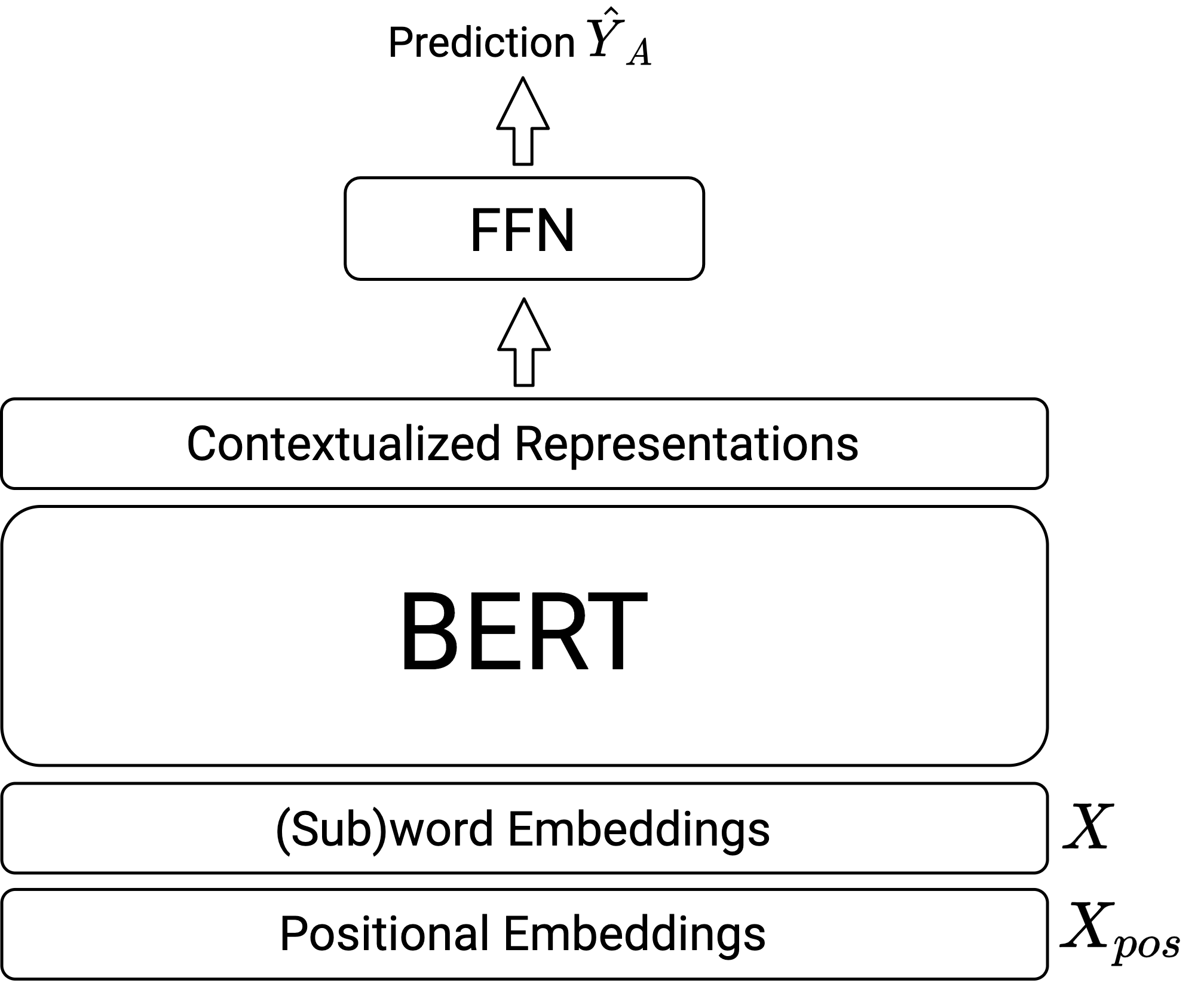}}
    \subfigure[Our MTL model]{\label{fig:mtl}\includegraphics[width=.45\linewidth]{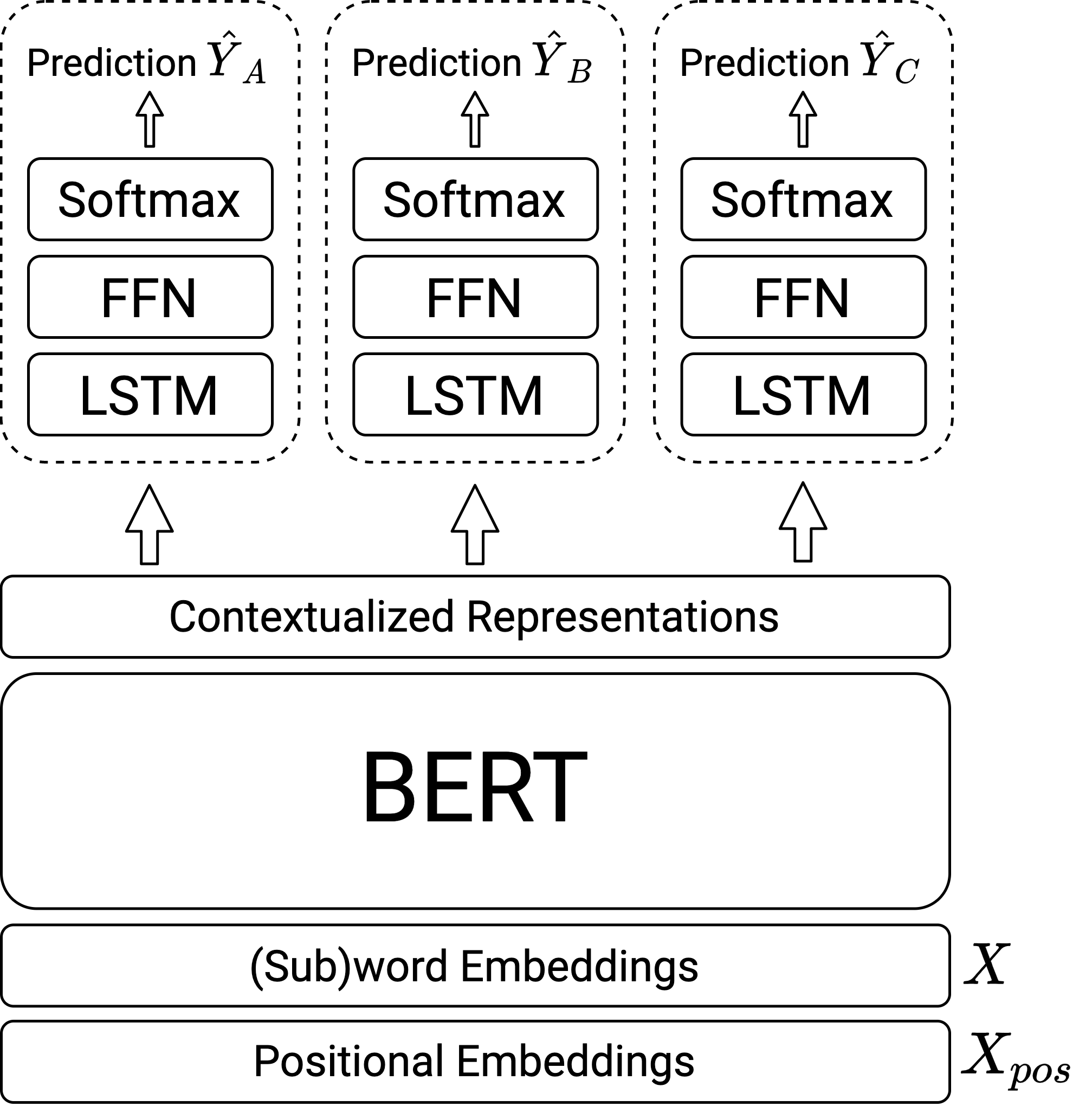}}
    \caption{(a) BERT baseline model. (b) Our MTL model. The bottom is a shared BERT backbone, and the upper parts are separate modules dedicated for each sub-task.}
\end{figure*}

\section{Methodology}
\label{sec:methodology}

We propose a Multi-Task Learning (MTL) method (Figure \ref{fig:mtl}) for this offensive language detection task. It takes good advantage of the nature of the OLID~\cite{zampieri-etal-2019-predicting}, and achieves an excellent result comparable to state-of-the-art performance only with the OLID \cite{zampieri-etal-2019-predicting} and no external data resources. A thorough analysis is provided in Section \ref{sec:result_analysis} to explain the reasons of not using the new SOLID dataset created in OffensEval-2020~\cite{zampieri-etal-2020-semeval}.


\subsection{Task Description}
The OffensEval-2020 \cite{zampieri-etal-2020-semeval} is a task that organized at SemEval-2020 Workshop. As mentioned in Section \ref{sec:dataset2}, it proposes a semi-supervised multilingual offensive language detection dataset which contains five different languages. It has three sub-tasks: (A) Offensive Language Detection; (B) Categorization of Offensive Language; (C) Offensive Language Target Identification. In this paper, we mainly focus on the sub-task A of the English data \cite{rosenthal2020}.


\subsection{Baseline}
\label{sec:baseline}

We re-implement the model of the best performing team~\cite{liu-etal-2019-nuli} in OffensEval-2019 \cite{zampieri-etal-2019-semeval} as our baseline. As illustrated in Figure \ref{fig:baseline}, \cite{liu-etal-2019-nuli} fine-tuned the pre-trained model, BERT~\cite{devlin2018bert}, by adding a linear layer on top of it.

\paragraph{BERT.} Bidirectional Encoder Representation from Transformer (BERT) \cite{devlin2018bert} is a large-scale masked language model based on the encoder of Transformer model \cite{transformer}. It is pre-trained on the BookCorpus \cite{bookcorpus} and English Wikipedia datasets using two unsupervised tasks:  (a) Masked Language Model (MLM) (b) Next Sentence Prediction (NSP). In MLM, 15\% of input tokens are masked, and the model is trained to recover them at the output. In NSP, two sentences are fed into the model and it is trained to predict whether the second sentence is the actual next sentence of the first one. As shown in \cite{devlin2018bert}, by fine-tuning, BERT achieves superior results on many NLP downstream tasks.

\subsection{Multi-task Offense Detection Model}
\label{sec:mtl}

In recent years, multi-task learning (MTL) technique is used in many machine learning fields to improve performance and generalization ability of a model \cite{Kang2011,Long015a,Kokkinos16,Riza2018,liu2018pad-net,dankers-etal-2019-modelling}. Generally, MTL has three advantages. Firstly, with multiple supervision signals, it can improve the quality of representation learning, because a good representation should have better performance on more tasks. Secondly, MTL can help the model generalize better because multiple tasks introduce more noises and prevent the model from over-fitting. Thirdly, sometimes it is hard to learn features by one task but easier to learn by another task. MTL provides complementary supervisions to one task and makes it possible to eavesdrop other tasks and get more information.

For this task, MTL is a very effective strategy. As mentioned in Section \ref{sec:dataset1} and shown in Table \ref{tab:performance-comparison}, the three labels in OLID are hierarchical and they are designed to be inclusive from top to bottom. This makes it possible for one sub-task to eavesdrop information form the other tasks. For example, if a tweet is labelled as Targeted in sub-task B, then it must be classified to Offensive in sub-task A.

Our MTL architecture is shown in Figure \ref{fig:mtl}. The bottom part is a BERT model, which is shared among all three sub-tasks. The upper parts are three separate modules dedicated for each sub-task, each module contains a Recurrent Neural Network (RNN) with Long-Short Term Memory (LSTM) cells \cite{hochreiter1997long}. The input \(X\) is first fed into the shared BERT, then each sub-task module takes the contextualized embeddings generated by BERT and produces a probability distribution for its own target labels. The overall loss \(L\) is calculated by \(L = \sum_i^I w_iL_i\). Here, $I = \{A, B, C\}$ and $w_i$ is the loss weight for each task-specific Cross-Entropy loss \(L_i\), where $\sum_i^I w_i = 1$. The loss weights are chosen by cross validation.



\section{Experiments}
\label{sec:experiments}

During the training phase, we evaluate our models on the test set of OLID (OffensEval-2019). As a reference, we also evaluate them on the test set of SOLID (OffensEval-2020), which is only released after the submission date.

\subsection{Experimental Settings}
\label{sec:exp_settings}
To find the optimal architecture for this task within the models we have, we set up five different experiments (Table \ref{tab:performance-comparison}). For the first two, we train our baseline model on OLID and SOLID separately. As SOLID's labels are AVG\_CONF scores between 0 to 1 rather than binary classes, we set the threshold as 0.3 to convert SOLID to a classification dataset. We set this threshold value by first training a BERT classifier on the OLID dataset, and then do predictions on SOLID and choose the best threshold by cross validation. Besides, we also conduct an experiment that pre-train the baseline model on SOLID and fine-tune on OLID by utilizing the pre-train strategy discussed in Session \ref{sec:analysis}. Finally, we train our Multi-task Offense Detection Model only on OLID and fine-tune the hyper-parameters based on Sub-task A (i.e. the goal is to leverage the information in level B and C to improve level A). To further improve the generalization performance of our method, we ensemble five MTL models with different parameter initialization and generate final results through majority voting.

To evaluate the performance of each model, we use macro-F1 which is computed as a simple arithmetic mean of per-class F1-scores. Since OLID released its test set last year, we use this test set as our validation set and optimize the hyper-parameters manually over the successive runs on it. For our best MTL model, we set the learning rate as 3e-6 and batch size as 32, the loss weights of subtasks A, B, C are 0.4, 0.3, 0.3 respectively. We train the model with maximum 20 epochs and utilize an early stop strategy to stop training if the validation macro-F1 doesn't increase in three continuous epochs. Our code is implemented in PyTorch and all experiments are run on a single GTX 1080Ti.


\subsection{Result Analysis}
\label{sec:analysis}
The results on Table \ref{tab:performance-comparison} show the macro-F1 scores on OLID and SOLID's test set and they are consistent except the model with pre-training. Our ensembled MTL model achieves the best performance in both two test sets.

\paragraph{Pre-train vs. No pre-train on SOLID.}
Since the SOLID \cite{zampieri-etal-2020-semeval} contains more than 9 million samples with the AVG\_CONF score. To make full use of the dataset, we conduct pre-train strategy which let the model pre-trained on SOLID and then fine-tuned on the Offensive Language Identification Dataset(OLID) \cite{zampieri-etal-2019-predicting}. To pre-train the model on SOLID, we regard the Sub-task A as a regression problem based on the AVG\_CONF score. Instead of setting a threshold to divide the data into two classes(OFF, NOT), we directly apply Mean Square Error(MSE) loss function on AVG\_CONF. However, our result shows that conducting pre-training makes little difference. We believe it is because the SOLID contains lots of noisy data which is also the reason why the baseline model trained on SOLID is much worse than on OLID.

\paragraph{BERT and Multi-Task Learning}
 From the result, we find that incorporating BERT and multi-task learning can help improve the macro-F1 score of Sub-task A a lot. This can be attributed to two reasons. Firstly, BERT model is pre-trained on a huge corpus which helps to produce more meaningful representations for the input text. Meanwhile, the large model size increases the learning ability for the task. Secondly, with the large capacity of BERT, through multi-task learning, sub-task A can get more information from the other shared part of the model, and it will be more certain to some cases. For example, if the label of sub-task B is \emph{NULL}, then label of sub-task A must be \emph{NOT}. If the label of sub-task B is \emph{TIN} or \emph{UNT}, then the label of sub-task A must be \emph{OFF}.
 
\begin{table}[!t]
\begin{center}
\begin{minipage}{0.8\textwidth}
\begin{tabular*}{\textwidth}{l @{\extracolsep{\fill}} cc}
\hline
\textbf{Model}                                                                         & \multicolumn{1}{c}{\textbf{F1 - OLID}} & \multicolumn{1}{c}{\textbf{F1 - SOLID}} \\ 
\hline
BERT (OLID)    & 0.8203            & 0.9088           \\
BERT (SOLID)    & 0.7280       & 0.9060           \\
BERT(Pre-trained with MSE loss)      & 0.8138       & 0.9107           \\
BERT + MTL     & 0.8341            & 0.9139           \\
BERT + MTL (Ensemble)    & \textbf{0.8382}       & \textbf{0.9151}           \\
\hline
\end{tabular*}
\end{minipage}
\end{center}
\caption{Experimental results on sub-task A. The evaluation metric is macro F1 score, which is official in OffensEval-2020.}
\label{tab:performance-comparison}
\end{table}
\label{sec:result_analysis}

\section{Conclusion and Future work}
\label{sec:conclusion}
From all of our experiments, we conclude that MTL improves the performance of Sub-task A in both OLID and SOLID. Moreover, our finding shows that pre-training Sub-task A as a regression task doesn't improve the model's performance. We think that there are several paths for further work. Firstly, more studies about the combination of the sub-tasks can be investigated for MTL.
This can show us more about the interaction between sub-tasks, and how much does one influence another.
Secondly, as mentioned in \cite{Kokkinos16}, simultaneously updating the model's parameters during MTL can have negative effects on optimization as the total gradients are too noisy. This becomes more significant when the number of tasks is large or the batch size is small. As a result, asynchronous optimizations for each task may provide a more stable gradient descent.

\section*{Acknowledgements}

This work is funded by MRP/055/18 of the Innovation Technology Commission, the Hong Kong SAR Government.


\bibliographystyle{coling}
\bibliography{coling2020}

\end{document}